\documentclass[11pt]{article}

\usepackage[a4paper,margin=1in]{geometry}
\usepackage{times}
\usepackage{microtype}

\usepackage{graphicx}
\usepackage{xcolor}
\usepackage{booktabs}
\usepackage{multirow}
\usepackage{array}

\usepackage{amsmath}
\usepackage{amssymb}
\usepackage{amsfonts}
\usepackage{amsthm}

\usepackage{algorithm}
\usepackage[noend]{algpseudocode}

\usepackage[numbers,sort&compress]{natbib}
\usepackage[hidelinks]{hyperref}
\usepackage[nameinlink,capitalise]{cleveref}

\theoremstyle{plain}

\theoremstyle{definition}

\theoremstyle{remark}

\newcommand{\ourmethod}{\emph{STRELGen}}
\newcommand{\ColSTREL}{Colored STREL}
\newcommand{\cSTREL}{CSTREL}

\newcommand{\until}{\mathtt{U}}
\newcommand{\ev}{\mathtt{F}}
\newcommand{\glob}{\mathtt{G}}

\newcommand{\reach}{\mathtt{R}}
\newcommand{\escape}{\mathtt{E}}
\newcommand{\surr}{\mathtt{Surr}}
\newcommand{\sw}{\mathtt{SW}}
\newcommand{\ew}{\mathtt{EW}}

\newcommand{\bb}{\mathbb}
\newcommand{\per}{\times}

\newcommand{\camera}[1]{#1}

\newtheorem{example}{Example}
\newtheorem{remark}{Remark}

\newcommand{\BibTeX}{\rm B\kern-.05em{\sc i\kern-.025em b}\kern-.08em\TeX}

\title{Guiding Neuro-Symbolic Scenario Generation with Spatio-Temporal Logic}

\author{
Lorenzo Bonin\thanks{Equal contribution.} \\
University of Trieste \\
\texttt{lorenzo.bonin@phd.units.it}
\and
Francesco Giacomarra\footnotemark[1] \\
University of Trieste \\
\texttt{francesco.giacomarra@phd.units.it}
\and
Luca Bortolussi \\
University of Trieste \\
\texttt{lbortolussi@units.it}
\and
Jyotirmoy V. Deshmukh \\
University of Southern California \\
\texttt{jyotirmoy.deshmukh@usc.edu}
\and
Francesca Cairoli \\
University of Trieste \\
\texttt{francesca.cairoli@units.it}
}

\date{}

\begin{document}

\maketitle

\begin{center}
\textit{Accepted at the 25th International Conference on Autonomous Agents and Multiagent Systems (AAMAS 2026).}
\end{center}

%%%%%%%%%%%%%%%%%%%%%%%%%%%%%%%%%%%%%%%%%%%%%%%%%%%%%%%%%%%%%%%%%%%%%%%%
% Abstract
%%%%%%%%%%%%%%%%%%%%%%%%%%%%%%%%%%%%%%%%%%%%%%%%%%%%%%%%%%%%%%%%%%%%%%%%

\begin{abstract}
The rapid advancement of autonomous driving (AD) technologies has outpaced the development of robust safety evaluation methods.
Conventional testing relies on exposing AD systems to vast numbers of real-world traffic scenes---a brute-force approach that is prohibitively expensive and statistically ineffective at capturing the rare, safety-critical edge cases essential for validating real-world robustness.
To address this fundamental limitation, we introduce \ourmethod, a scalable framework for the targeted generation of safety-critical driving scenarios. \ourmethod\ synergistically combines a multi-agent trajectory-generation diffusion model (DM) with Spatio-Temporal Logic (STREL) specifications that encode complex safety and realism properties through a highly interpretable formalism. Crucially, monitoring satisfaction levels of these specifications is differentiable, enabling gradient-based search. At inference time, we optimize directly over the DM's latent space to maximize STREL formula satisfaction.
The result is efficient generation of highly plausible yet safety-critical multi-agent scenarios that lie within the learned data distribution. \ourmethod\ thus provides a flexible, interpretable, and powerful tool for stress-testing autonomous driving systems, moving beyond the limitations of brute-force data collection.

%\fc{While autonomous driving (AD) technologies are advancing rapidly, the development of robust safety evaluation frameworks remains a critical unsolved problem. Conventional testing seeks to validate safety by exposing AD systems to a vast number of driving scenes, a brute-force approach that is as expensive as it is ineffective at capturing critical rare events. However, these very events -- the edge cases at the distributional tail -- are precisely what is needed to prove real-world robustness. To overcome these limitations, we introduce a scalable framework that leverages diffusion models (DMs) and spatio-temporal logic (STREL) to systematically generate high-quality, safety-critical driving scenarios. Our \ourmethod\ framework is built around a DM trained on the Argoverse 2 dataset to generate realistic, multi-agent trajectories from an initial scene. To steer this generation toward safety-critical scenarios, we employ STREL specifications that encode complex safety and realism properties. Monitoring the quantitative satisfaction of such requirements is a differentiable task allowing for gradient-based searches. At inference time, we directly optimize over the DM's latent space to maximize the satisfaction of these STREL formulas, ensuring the generated scenarios are both safety-critical and plausible, i.e. in-distribution.

%\ourmethod\ offers a flexible and interpretable data augmentation to test AD technologies against.} Experiments ... 
\end{abstract}

%%%%%%%%%%%%%%%%%%%%%%%%%%%%%%%%%%%%%%%%%%%%%%%%%%%%%%%%%%%%%%%%%%%%%%%%
% Main paper
%%%%%%%%%%%%%%%%%%%%%%%%%%%%%%%%%%%%%%%%%%%%%%%%%%%%%%%%%%%%%%%%%%%%%%%%

\section{Introduction}\label{sec:intro}

The integration of autonomous vehicles (AVs) into everyday transportation is progressing rapidly, but this trend is accompanied by growing concerns regarding safety. Developing self-driving systems capable of ensuring reliable and secure operation represents a multifaceted and technically demanding challenge.
%%% LORENZO's
%Recent advances in machine learning based approaches have enabled autonomous driving systems to achieve high levels of reliability and robustness across diverse driving scenarios, with large-scale deployments such as Waymo's fleet proving human-competitive performance~\cite{di2024comparative}. 
%%%%
Machine learning (ML) has enabled autonomous driving systems to achieve human-competitive reliability and robustness across diverse scenarios, a milestone demonstrated by large-scale deployments like the Waymo fleet~\cite{di2024comparative}. However, designing rigorous and comprehensive testing environments remains a major difficulty, particularly when it comes to exposing the system to rare but safety-critical scenarios essential for robust evaluation. 
Current approaches rely on large-scale real-world deployments and simulator-based replay of safety-critical events~\cite{webb2020waymo}. Nonetheless, the enormous search space, intricate interactions, and rarity of critical events make autonomous vehicle safety testing both highly challenging and often ineffective~\cite{liu2024curse, song2023critical}.
% Data-driven simulators typically replay or modify recorded driving logs, limiting exploration due to the finite diversity of available datasets.
% Since validating AV safety requires billions of miles of diverse simulation, log-based approaches lack the scalability for reliable training and evaluation. Importantly, AV environments can be modeled as multi-agent systems, where each test {\em scenario} represents a set of physically realistic trajectories within this system.
Log-based simulators are limited by finite dataset diversity, hindering scalability for reliable training and evaluation. %Importantly, AV environments can be viewed as multi-agent systems, with each test {\em scenario} representing a set of physically realistic trajectories within this system.

In general, AV environments can be viewed as multi-agent systems, with each test {\em scenario} representing a set of physically realistic trajectories within this system.
With recent advances in deep generative models, particularly diffusion models (DMs)~\cite{sohldickstein2015_diffusion, ho2020_ddpm}, a promising approach is to generate scenarios directly, rather than relying solely on samples from real-world data~\cite{pronovost2023_scenariodiffusion}. In this context, DMs can be used to generate trajectories for autonomous agents, typically conditioned on contextual information such as road maps, traffic configurations, or environmental settings to ensure consistency with scene dynamics. However, without a dedicated mechanism to control the generation process after the training phase, the objective remains limited to producing realistic trajectories under the given context, without explicitly steering the model toward desired behaviors. Consequently, evaluating the system under specific, potentially rare or challenging behaviors---those unlikely to arise in typical trajectories---remains an open problem. %becomes problematic. 
To address this, other works~\cite{scassola2025zeroshot,huang2024_versatile, xu2025_diffscene} %have proposed incorporating 
incorporate guidance mechanisms to elicit desired properties. Such approaches achieve controllable generation by introducing guidance at each step of the denoising process of the DM, where network outputs are perturbed using the gradient of a differentiable objective to promote desired behaviors. When applied to traffic simulation, however, designing and implementing such objectives---e.g., adherence to traffic rules, safety distance, realism---becomes highly challenging due to the inherently spatio-temporal and multi-agent characteristics of the domain. To tackle this challenge, Zhong et al.~\cite{zhong2022guidedconditionaldiffusioncontrollable} proposed leveraging the established syntax of Signal Temporal Logic (STL)~\cite{maler2004_stl}. As a formal language designed for specifying %spatio-temporal 
temporal constraints, STL provides a systematic and scalable framework for defining driving rules. In addition, it incorporates a quantitative notion of robustness, allowing the degree to which rules are satisfied to be formally measured. While STL is well-suited for specifying properties for a single agent or a small, fixed pool, it faces significant scalability challenges in multi-agent systems. The core issue is that specifying the complex interactions between agents exponentially increases the complexity and size of the STL formulae. This combinatorial explosion not only affects computational performance but, more critically, severely compromises the interpretability of the requirements. Consequently, writing exhaustive and correct specifications for collaborative or competitive behaviors becomes an inherently hard and error-prone task.

In this work, we extend controllable trajectory generation by leveraging Signal Spatio-Temporal Reach and Escape Logic (STREL)~\cite{bartocci2017_monitoring,nenzi2022logic}, which generalizes STL by incorporating explicit spatial relationships between agents and the environment. At any given timestep, the scene is modeled as a graph where nodes represent agents (e.g., cars, pedestrians). Each node is characterized by a set of attributes, including categorical features like agent type and continuous states like position and velocity. Compared to STL, STREL enables more expressive specifications that capture both temporal and spatial dependencies, allowing fine-grained control over realistic traffic scenarios. A significant limitation of STREL is its difficulty in monitoring the quantitative satisfaction of properties involving categorical predicates. To overcome this, we introduce Colored STREL, an extension of STREL that enables the specification of predicates over specific node types.

Our approach partitions the spatial graph into color-specific subgraphs, where colors represent categorical attributes (e.g., vehicle type in autonomous driving scenes). This allows Colored STREL to monitor spatial properties for particular agent classes and, crucially, to derive continuous robustness values even from discrete categorical attributes. This capability facilitates the precise enforcement of complex, interpretable behaviors in generated trajectories.

Furthermore, we leverage Colored STREL for guided trajectory generation. By formulating specifications as objective functions, we can steer a generative model toward high-satisfaction outputs. This guidance is fully differentiable, enabling efficient gradient-based optimization to identify latent inputs that produce trajectories satisfying our requirements, all without the need for model retraining. Overall, our approach enables controllable trajectory generation, ensuring that the generated scenarios adhere to clearly defined, semantically meaningful specifications.

The \textit{main contributions} presented in this paper can be summarized as follows:

\begin{itemize}
    \item[\textbf{1.}] \textbf{\ColSTREL: A Colored Spatio-Temporal Logic.} We introduce \ColSTREL, a novel extension of STREL that partitions spatial graphs by node attributes (e.g., vehicle type). This enables the monitoring of complex spatial properties across specific classes of objects while maintaining continuous robustness semantics. The resulting monitor is fully differentiable.
    \item[\textbf{2.}] \textbf{{\ColSTREL} Guidance.} We present a differentiable guidance framework that uses {\ColSTREL} specifications as objective functions. This allows us to directly optimize the latent space of generative models to produce trajectories that satisfy spatio-temporal requirements.
    \item[\textbf{3.}]\textbf{Safety-Critical Data Augmentation.} We present a data augmentation strategy that uses our guidance framework to generate diverse, realistic, and safety-critical scenarios that are underrepresented in real-world data, ensuring consistency with traffic rules and preventing unrealistic outputs.
\end{itemize}

\subsection*{Related Work}\label{subsec:rel_works}

\paragraph{Scenario Generation with Diffusion Models}
Several works have leveraged diffusion models to synthesize adversarial or risky driving behaviors targeting the ego vehicle. Xu et al.~\cite{xu2025_diffscene} proposed DiffScene, where a diffusion model is first trained to generate goal-agnostic trajectories of surrounding vehicles. At inference time, adversarial guidance is applied by optimizing a composite risk objective, which balances safety (increasing ego-vehicle risk), functionality (maintaining task feasibility), and realism. While effective, their experiments primarily focus on single surrounding vehicles, limiting scalability to complex multi-agent settings.

Other approaches emphasize the generation of realistic multi-agent interactions. Pronovost et al.~\cite{pronovost2023_scenariodiffusion} introduced a latent diffusion model conditioned on map-based Bird's-Eye-View representations and structured token descriptions to generate realistic poses and trajectories of multiple agents in the Argoverse 2 dataset. Huang et al.~\cite{huang2024_versatile} proposed Versatile Behavior Diffusion, which integrates Transformer-based encoders and denoisers with classifier guidance to generate joint behaviors of multiple traffic agents. Rowe et al.~\cite{rowe2025_scenariodreamer} presented Scenario Dreamer, a vectorized latent diffusion approach that separately generates initial traffic scenes and closed-loop behaviors, and supports exponential tilting to bias scenarios towards adversarial or benign outcomes. 
These works highlight the growing interest in scaling diffusion-based scenario generation to complex, interactive traffic environments.

\paragraph{Temporal Logic for Autonomous Driving}
 A line of research has explored the integration of Signal Temporal Logic (STL) into the autonomous driving domain. The techniques in~\cite{gressenbuch2021_predictive,cairoli2023conformal,lindemann2023conformal,cairoli2025genqpm} explicitly predict traffic rule violations by combining temporal logic with neural networks. In these frameworks, traffic rules are formalized in STL, and robustness values are computed to serve as features for learning models that anticipate violations in highway driving scenarios. Results show that these predictive monitoring approaches can outperform conventional methods that first predict trajectories and then check for compliance, highlighting the potential of temporal logic as a supervisory signal.

In parallel, NVIDIA researchers introduced a framework that uses STL-based guidance to generate and evaluate safety-critical driving scenarios ~\cite{zhong2022guidedconditionaldiffusioncontrollable}. Their philosophy is similar to ours in that STL formulae are used to encode safety properties and realism constraints for autonomous vehicle testing. However, their approach relies on  gradient-guided optimization at generation time, whereas our method performs latent-space search using STREL, which is more well-fitted for spatial domains. This distinction is crucial, as it allows us not only to target safety-critical scenarios but also to systematically assess the coverage of the generative model with respect to rare events, something not addressed in prior works.

\paragraph{Noise-Optimization Guidance Approaches}
Related to our proposed guidance approach, there exists a growing line of works that explore the possibility of steering the generation process from a pre-trained diffusion model by optimizing the noise vector in the latent space, according to some differentiable objective function defined on the output space. 
 DOODL  \cite{wallace2023doodl} and Direct Noise Optimization (DNO) \cite{karunratanakul2024dno} exemplify this trend, as they both present a guidance approach based on back-propagating through the entire reverse process to find latent points that yield outputs best aligned with a target classifier or motion objective. A similar approach to what we propose can be found in \cite{giacomarra2025certified}, where robustness with respect to STL formulae is used to guide an optimization process in the latent space of score-based diffusion models.
 
 Such methods enable fine-grained, differentiable control over generation without retraining, but require careful regularization to avoid drifting away from the data manifold.

%%%%%%%%%%%%%%%%%%%%%%%%%%%%%%%%%%%%%%%%%%%%%%%%%%%%%%%%%%%%%%%%%%%%%%%%

%\todo{STRUCTURE OF THE PAPER}

\section{Background}

In this section, we review the theoretical foundations underlying our approach.  %, focusing on diffusion models for trajectory generation and the formal framework of spatio-temporal logic used for behavior specification and guidance.

\subsection{Diffusion Models}
\label{sec:considered_diffusion}

Diffusion probabilistic models~\cite{sohldickstein2015_diffusion} are generative models that learn data distributions by gradually adding and then removing noise from training samples. We consider the denoising diffusion probabilistic model (DDPM)~\cite{ho2020_ddpm}, which can be viewed as a discretized score-based model trained with Langevin dynamics~\cite{song2020_sde}.  

\paragraph{Forward and Reverse Processes.}
The forward process progressively perturbs a clean sample $x^0$ through a sequence of $\mathcal{T}$ injections of Gaussian noise with variances $\beta_\tau$, $\tau\in\{1,\dots,\mathcal{T}\}$, until it reaches a pure noise distribution $x^\mathcal{T} \sim \mathcal{N}(\vec{0}, I)$. The reverse process, parameterized by a neural denoiser $\epsilon_\theta(x^\tau,\tau)$, learns to invert this corruption, reconstructing clean samples from noisy ones. The model is trained via a simple denoising objective that minimizes the prediction error between true and estimated noise:
\begin{equation}\label{eq:diff_obj}
\mathcal{L}_{\text{diff}}(\theta)
= \bb{E}_{\tau, x_0}\!\left[\|\epsilon - \epsilon_\theta(x^\tau,\tau)\|^2\right],
\end{equation}
where $\it \tau\sim Unif(\{1,\dots,\mathcal{T}\})$ and $x_0$ is sampled from the dataset. 
The model can then generate new samples by iteratively denoising pure noise drawn from $ \mathcal{N}(\vec{0}, I)$.

\paragraph{Conditional Diffusion.}
To guide generation, diffusion models can be conditioned on auxiliary information $y$ (e.g., scene context or class labels), learning the conditional distribution $p(x^0 \mid y)$. The same denoising loss applies, with the denoiser now depending on $y$, i.e., $\epsilon_\theta(x^\tau,\tau \mid y)$. This enables targeted generation of trajectories consistent with external constraints.

\paragraph{Implicit and Latent Variants.}
%Sampling from standard DDPMs can be slow due to the need to go through many stochastic denoising steps.  
The sampling process in DDPMs is slow, as it requires executing a long chain of probabilistic denoising steps.
Denoising Diffusion Implicit Models (DDIMs) \cite{song2022_ddim} allow for faster sampling by removing stochasticity from all denoising steps except the last one (step $\mathcal{T}$). Therefore, given a sample $x^\mathcal{T} \sim \mathcal{N}(\vec{0}, I)$, the rest of the denoising process becomes deterministic. The sampling process can be then accelerated by using only a subset of the diffusion steps. %by skipping denoising steps and 
%yielding deterministic and differentiable trajectories. %while preserving the same training objective.  
Finally, Latent Diffusion Models (LDMs)~\cite{rombach2022high} perform the diffusion process in a compressed latent space learned by an autoencoder, greatly improving efficiency while maintaining generative performances.

\subsection{Spatio-Temporal Logic}\label{subsec:strel}

\begin{figure}[!t]
    \centering
\includegraphics[width=.8\columnwidth]{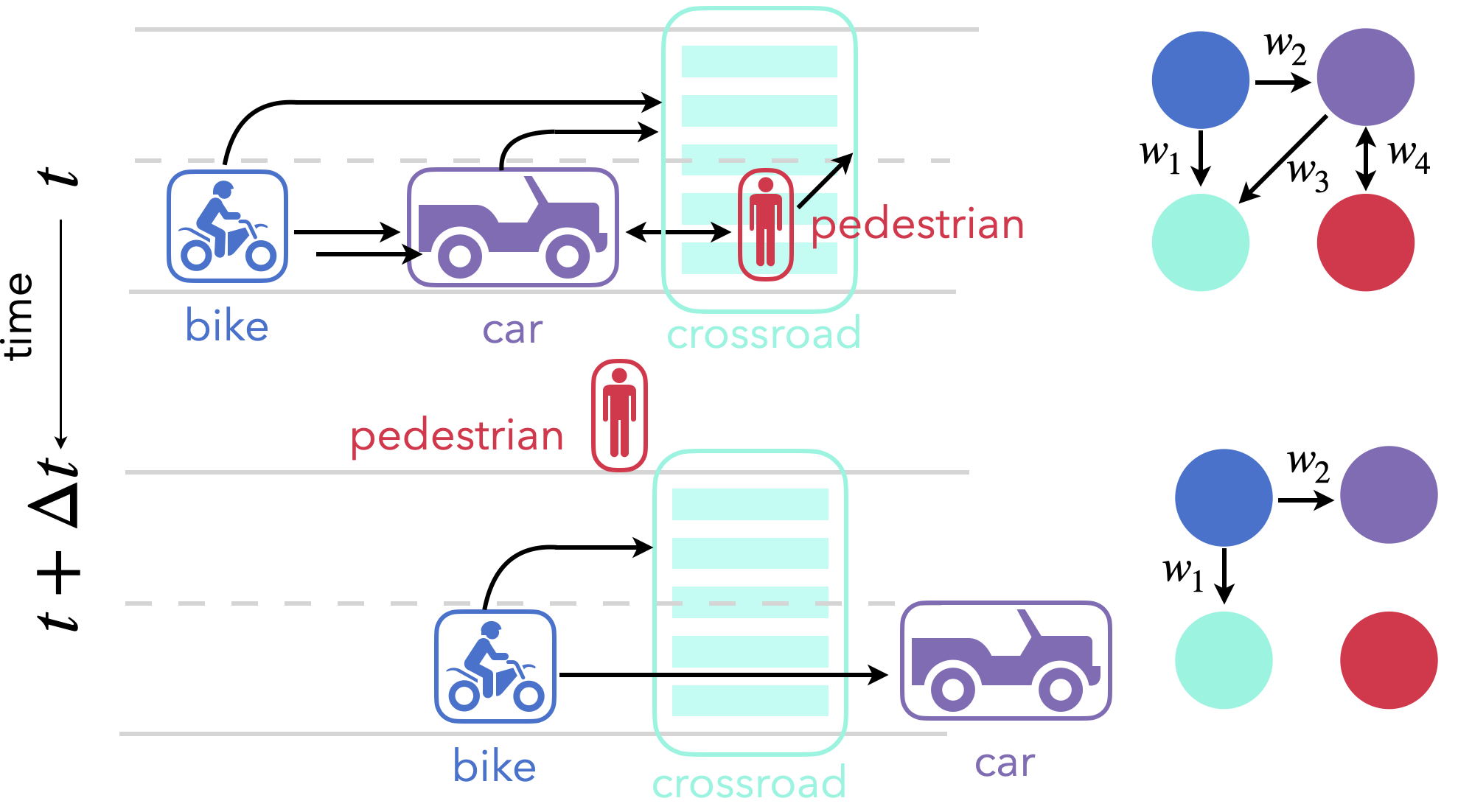}
    \caption{Example of a safety-critical AD scene represented as a STREL dynamic graph: the motorbike sees the crossroad but not the pedestrian, as it is hidden by the car in front.}
    
    \label{fig:running1}
\end{figure}

Spatio-Temporal Reach and Escape Logic (STREL)~\cite{bartocci2017_monitoring,nenzi2022logic} is a formal language designed to express complex, spatio-temporal relationships between interacting agents. In this framework, a multi-agent system is modeled as a dynamic graph $G(t) = (L(t), E(t))$, where $L(t)$ is the set of agents (nodes) and $E(t)$ represents their possible interactions (edges) at time $t$. Each edge has two key properties: a binary weight indicating whether two agents are interacting (e.g., if they see each other), and a distance metric 
$f$ that quantifies a specific notion of proximity between them. This metric $f$ is a key source of STREL's expressivity, allowing it to define a wide range of spatial properties. Crucially, the graph is dynamic; the set of agents, their attributes, and the connections between them all evolve over time, as illustrated in Fig.~\ref{fig:running1}.

STREL properties are defined by the following \textit{syntax}:
\begin{equation*}
    \varphi:= true\mid\mu\mid\neg\varphi\mid \varphi_1\land\varphi_2\mid \varphi_1 \until_{[t_1,t_2]}\varphi_2\mid
    \varphi_1 \reach_{[d_1,d_2]}^{f}\varphi_2\mid \escape_{[d_1,d_2]}^{f}\varphi 
\end{equation*}
where $\mu$ denotes the atomic predicates, $\neg$, $\land$ denote negation and disjunction. On one hand we have the temporal operator \textit{Until}, $\varphi_1 \until_{[t_1,t_2]}\varphi_2$, that denotes that $\varphi_1$ is satisfied until, in a time between $t_1$ and $t_2$ time units in the future, $\varphi_2$ becomes true. Temporal operators are evaluated in each location separately. 
On the other hand, $\reach_{[d_1,d_2]}^{f}\varphi_2$ and $\escape_{[d_1,d_2]}^{f}\varphi$ denote the spatial operators of \textit{Reach} and \textit{Escape}, with $f:L\per L\to B$ a distance function ($B$ the distance domain). % that computes the weight associated to associate to edges between nodes. 
The reachability operator describes the property of reaching a location
satisfying $\varphi_2$, through a path with all locations that satisfy $\varphi_1$ with path length between $d_1$ and $d_2$. The escape operator describes the possibility of escaping from a certain region via a path passing only through locations that satisfy $\varphi$, with the distance between the starting location of the path and the last (not the path length) that belongs to the interval $[d_1,d_2]$. Spatial operators are evaluated at each time step separately. 
From this essential syntax, we can define as usual other operators as follows: $false:=\neg true$, $\varphi \lor \psi := \neg(\neg\varphi\land\neg\psi)$, $\ev_{[t_1,t_2]} \varphi:= true\ \until_{[t_1,t_2]}\varphi$ and $\glob_{[t_1,t_2]} \varphi:= \neg\ev_{[t_1,t_2]}\neg\varphi$, where $\ev_{[t_1,t_2]}\varphi$ and $\glob_{[t_1,t_2]}\varphi$ denote respectively the \emph{Eventually} ($\varphi$ will hold at some point in the time interval $[t_1,t_2]$) and \emph{Globally} ($\varphi$ holds at all times in the interval $[t_1,t_2]$) operator.

We can also derive other three spatial operators: 
\begin{itemize}
    \item {\bf somewhere}: $\sw^f_{[0,d]}\varphi:=true\reach^f_{[0,d]}\varphi$,
    \item {\bf everywhere}: $\ew^f_{[0,d]}\varphi:=\neg\sw^f_{[0,d]}\neg\varphi$,
    \item {\bf surround}: $\varphi_1\surr^f_{[0,d]}\varphi_2:= \varphi_1\land \neg\big(\varphi_1\reach_{[0,d]}\neg(\varphi_1\lor\varphi_2)\big)\land\neg(\escape_{[d,\infty]}\varphi_1)$.
\end{itemize}
\textit{Somewhere} and \textit{Everywhere} operators describe behaviour for some or all locations within reach of a specific location, \textit{Surround} expresses the notion of being surrounded by a region that satisfies $\varphi_2$, while being in a $\varphi_1$ satisfying region.

\paragraph{STREL Semantics}
The semantics of STREL is evaluated point-wise at each time $t$ and at each
location $\ell$.
%The satisfaction of a formula $\varphi$ by a signal $\vec{x}$ at time $t$ is defined by:
Let $G$ be a spatial model (i.e. a graph with a time-varying edge relation) with $L$ the space universe (i.e. the set of locations), $D_1$ and $D_2$ be two signal domains, and $\vec{x}$ be a spatio-temporal $D_1$-trace for locations in $L$. The
$D_2$-monitoring function $\mathbf{m}$ of $\vec{x}$ is recursively defined as follows.
\begin{align*}
\mathbf{m}(G,\vec{x},&\mu,t,\ell) = \
g(\mu,\vec{x}(\ell, t))\\
\mathbf{m}(G,\vec{x},&\neg\varphi,t,\ell) =  \odot \mathbf{m}(G,\vec{x},\varphi,t,\ell)\\
\mathbf{m}(G,\vec{x},&\varphi_1\land\varphi_2,t,\ell) = \mathbf{m}(G,\vec{x},\varphi_1,t,\ell) \otimes \mathbf{m}(G,\vec{x},\varphi_2,t,\ell)\\
\mathbf{m}(G,\vec{x},&\varphi_1 \until_{[t_1,t_2]}\varphi_2,t,\ell) =\bigoplus_{ t'\in [t+t_1,t+t_2]} \\
& \Big[\mathbf{m}(G,\vec{x},\varphi_2,t',\ell)  \otimes
\Big(\bigotimes_{t"\in[t,t']} \mathbf{m}(G,\vec{x},\varphi_2,t",\ell)\Big)\Big]\\
%\text{SINCE: } &\mathbf{m}(G,\vec{x},\varphi_1 \mathcal{S}_{[t_1,t_2]}\varphi_2,t,\ell) =\bigoplus_{t'\in [t-t_2,t-t_1]}\Big[\\
%&\mathbf{m}(G,\vec{x},\varphi_2,t',\ell)\otimes\Big(\bigotimes_{t"\in[t',t]} \mathbf{m}(G,\vec{x},\varphi_1,t",\ell)\big)\Big)\Big]\\
\mathbf{m}(G,\vec{x},&\varphi_1 \reach_{[d_1,d_2]}^{f}\varphi_2,t,\ell) =\bigoplus_{\tau\in Routes(G(t),\ell)}\Big[ \bigoplus_{i:\big(d_\tau^f[i]\in[d_1,d_2]\big)}\\
&\Big(\mathbf{m}(G,\vec{x},\varphi_2,t,\tau[i])\otimes\bigotimes_{j<i}\mathbf{m}(G,\vec{x},\varphi_1,t,\tau[j])\ \Big)\ \Big]\\
\mathbf{m}(G,\vec{x},&\escape_{[d_1,d_2]}^{f}\varphi,t,\ell) = \bigoplus_{\tau\in Routes(G(t),\ell)}\Big[\bigoplus_{\ell'\in\tau:\left(d_{G(t)}^f[\ell,\ell']\in[d_1,d_2]\right)}\\
&\Big( \bigotimes_{i\le\tau(\ell')}\mathbf{m}(G,\vec{x},\varphi,t,\tau[i])\ \Big)\ \Big]
\end{align*}
%$Routes(G(t))$ denotes the set of routes in $G(t)$, while 
Here, $Routes(G(t), \ell)$ denotes the set of routes in $G(t)$ starting from $\ell\in L$.
To ease the notation we use $\otimes, \oplus, \odot$ to denote respectively \camera{conjunction} $\otimes_{D_2}$, \camera{disjunction} $\oplus_{D_2}$ and \camera{negation} $\odot_{D_2}$. For the Boolean signal domain ($D_2 =  \{\bot,\top\}$),
we say that $(G , \vec{x}(\ell, t))$ satisfies a formula $\varphi$ iff $\mathbf{m}(G , \vec{x}, \varphi, t, \ell) = \top$. For max/min signal domain ($D_2=\bb{R}^\infty$) we say that $(G , \vec{x}(\ell, t))$ satisfies a formula $\varphi$ iff $\mathbf{m}(G , \vec{x}, \varphi, t, \ell) > 0$. 
%The \textit{Reach} operator describes the behaviour of reaching locations satisfying $\varphi_2$ through a path $\tau$ such that $d_1\le d^f(\tau) \le d_2$ and $\varphi_1$ is satisfied at all locations in the path. The \textit{Escape} operator describes the possibility of escaping from a certain region via a path that passes only through locations that satisfy $\varphi_1$ such that the distance between the starting location and the last location belongs in the interval $[d_1,d_2]$. Here, the distance constraint is not necessarily over the satisfying path, but the shortest distance path between the start and end location.

\begin{example} \label{ex:running1}
Fig.~\ref{fig:running1} illustrates the evolution in time of a safety-critical autonomous driving scene and its corresponding abstraction as STREL dynamic graph. In this model, a directed edge from node $a$ to node $b$ exists if and only if agent $a$ perceives agent $b$. %(i.e., $f(a,b)>0$). 
Crucially, the graph structure depends on the chosen distance function $f$, as different functions can model different perceptual or spatial relationships. In our example, the function emulates a LiDAR such that the motorbike perceives the crosswalk and the car in front, but not the occluded pedestrian behind it, accurately reflecting the scene's visibility constraints.
A potential safety-critical situation occurs if the motorbike finds a pedestrian ahead within the safety distance $d_{\it safe}$. This is captured by the following STREL requirement:
\begin{equation}\label{eq:runn1}
    \varphi = \ev_{[0,T]}\big[ \big(\mathtt{isMotorBike}(\vec{x})\big)\reach_{[0,d_{\it safe}]}^{Front}\big(\mathtt{isPedestrian}(\vec{x})\big)\big],
\end{equation}
where $\mathtt{isMotorBike}(\vec{x})$ 
and $\mathtt{isPedestrian}(\vec{x})$ are two atomic predicates denoting whether the node is of type motorbike or pedestrian. %, respectively.
\end{example}

This very simple example highlights a big \textit{limitation of STREL}. We are only able to monitor the Boolean satisfaction of $\varphi$, whereas the quantitative semantics is not well-defined over categorical attributes; we are not able to quantify how close the scene is to a safety violation. Therefore, the satisfaction of STREL formulae is unsuitable as an objective function for optimization problems involving categorical data, such as vehicle type.

\section{Neurosymbolic Scenario Generation}
In this section, we present our {\ourmethod} approach for autonomous driving scenario generation. The primary objective is to enable controllable generation of multi-agent trajectories that are both realistic and consistent with semantically meaningful specifications. To this end, we combine latent DMs for trajectory synthesis with \cSTREL-based guidance that steers the generative process toward desired behaviors.

Our method introduces two key innovations. First, we extend STREL with a coloring mechanism, yielding \emph{Colored STREL} (CSTREL), which partitions the spatial graph into subgraphs corresponding to agent types or contextual categories. This extension enables rules to be specified over heterogeneous entities (e.g., cars, pedestrians, cyclists) and provides continuous robustness measures even for discrete categorical attributes. Second, we develop a differentiable guidance mechanism that directly leverages {\cSTREL} robustness values as objectives for trajectory generation. By expressing logical constraints as differentiable functions, this mechanism enables gradient-based search in the latent space, ensuring that generated trajectories satisfy the specified constraints.

% An overview of the proposed approach is illustrated in Figure\textcolor{red}{[add Figure]}.

\subsection{Trajectories Generation}

In this work, we adopt the diffusion-based architecture proposed in~\cite{wang2024_optimizingdiffusion} as our baseline for multi-agent scenario synthesis. 
% The model is trained on the Argoverse 2 dataset~\cite{wilson2023_argoverse2} using the same hyperparameter settings reported in their study. 
A latent diffusion model is used to generate future trajectories of all agents in a traffic scene jointly, while scene context---comprising map features, agent history, and inter-agent interactions---is encoded by a pre-trained QCNet~\cite{zhou2023_qcnet} and injected into the diffusion model through cross-attention layers. This conditioning ensures that the generated trajectories remain consistent with both scene geometry and agent dynamics.
\camera{A central contribution of~\cite{wang2024_optimizingdiffusion} %their work 
is the introduction of Optimal Gaussian Diffusion (OGD), which replaces the standard Gaussian prior with a data-dependent “optimal” prior derived from %marginal trajectory statistics at a small diffusion time $T$. 
statistics of marginal trajectories at a small diffusion time step.}
This design allows the model to achieve high-quality predictions with substantially fewer denoising steps. 
% In addition, Wang et al. propose Estimated Clean Manifold Guidance (ECM) and its variant ECMR, which address the inefficiency of traditional guided sampling. Rather than injecting gradient penalties at noisy intermediate diffusion states, ECM shifts guidance to the clean data manifold, while ECMR further leverages warm starts from reference trajectories. Both techniques reduce computational cost without sacrificing guidance effectiveness.

In our framework, we adopt the implicit formulation of~\cite{wang2024_optimizingdiffusion}, which accelerates sampling and, crucially, renders the reverse process deterministic. This property is particularly important for our method, as it enables stable backpropagation of gradient-based {\cSTREL} specifications defined on generated trajectories with respect to latent inputs.

\subsection{Colored STREL}
Scenarios can be naturally interpreted as networks of interacting agents. Spatio-temporal logic \camera{(STREL in
Section~\ref{subsec:strel})} is tailored to monitor dynamic networks of spatially-distributed agents. A key feature is the %possibility of
\camera{ability to} automatically monitor %the
\camera{both Boolean and quantitative} 
satisfaction of spatial and temporal properties, %both in its Boolean and quantitative implementation, 
%allowing for an understanding of complex behaviours that can emerge from local and dynamic interactions. 
\camera{providing insights into complex behaviors that emerge from local and dynamic interactions.}
%We extend STREL (presented in  Section~\ref{subsec:strel}) to its a colored formulation, referred to as {\ColSTREL} (\cSTREL). %The need for this extension is motivated by the limitation depicted in Example~\ref{ex:running1} showing that the
%quantitative semantics is not well-defined over properties that contain categorical atomic predicates, identifying whether a node is of a certain type. Our interest is to be able to discern between properties that must hold, for instance, for cars or pedestrians, without losing the quantitative semantics, which is an essential metric in our guidance framework. 
\camera{However, }the STREL quantitative semantics---a core metric in our guidance framework---is currently ill-defined for properties containing categorical predicates (e.g., $\mathtt{isBike}$ or $\mathtt{isPedestrian}$), as shown in Example~\ref{ex:running1}. \camera{To address this, we introduce a colored formulation, {\ColSTREL} (\cSTREL), which enables the specification and quantitative verification of type-specific requirements.} %To address this, we propose an extension that preserves the essential quantitative metrics of our framework while allowing us to specify and verify type-specific requirements.
Given the set of all colors $C$, e.g. $\it C=\{\texttt{car}, \texttt{bike}, \texttt{pedestrian}, \texttt{traffic light}, \texttt{crossroad}\}$, the syntax of {\ColSTREL} is:
\begin{equation*}
    true\mid\mu^c\mid\neg\varphi\mid \varphi_1\land\varphi_2\mid \varphi_1 \until_{[t_1,t_2]}\varphi_2\mid
\varphi_1^{c_1}\reach_{[d_1,d_2]}^{f}\varphi_2^{c_2}\mid \escape_{[d_1,d_2]}^{f}\varphi^{c} ,
\end{equation*}
where $c, c_1, c_2\subseteq C$ are sets of colors. If $c = c_1 = c_2 = C$, {\cSTREL} reduces exactly to STREL. 
To better understand the main differences in the colored semantics, let's introduce a \textit{coloring function}, $p:L\per \bb{R}^+\to C$, mapping every pair $(\ell,t)$, node $\ell\in L$ at time $t\in \bb{R}^+$, into the associated color $p(\ell,t)\in C$.

Let $\perp_{D_2}$ denote the minimum satisfaction value, i.e. $False$ in the Boolean case and $-\infty$ in the quantitative case. The main differences in {\cSTREL} semantics can be summarized as follows.
%\begin{align*}
%   \textbf{Colored Atoms:    } \mathbf{m}(S,\vec{x},\mu^c,t,\ell) = \
%g_c(\mu^c,\vec{x}(\ell, t)),
%\end{align*}
$$\hspace{-1.5cm}\textbf{Colored Atoms: }\qquad \mathbf{m}(S,\vec{x},\mu^c,t,\ell) = g_c(\mu^c,\vec{x}(\ell, t)),$$
 where $g_c(\mu,\vec{x}(\ell, t)) = g(\mu,\vec{x}(\ell, t))$ if $p(\ell,t)\in c$, i.e., if node $\ell$ at time $t$ has a color present in set $c$, and $g_c(\mu,\vec{x}(\ell, t)) = \perp_{D_2}$ otherwise.

For the colored reach and escape operators, we must address the path-generating function in order to search only over paths with allowed coloring. A collateral advantage of {\cSTREL} is its reduced computational cost, as we considerably reduce the search space.
\begin{align*}
\textbf{C}&\textbf{olored Reach:}\quad 
\mathbf{m}(S,\vec{x},\varphi_1^{c_1} \reach_{[d_1,d_2]}^{f}\varphi_2^{c_2},t,\ell) =\bigoplus_{\tau\in CRoutes(S(t),\ell,c_1,c_2)}\\ &\Big[ \bigoplus_{i:\big(d_\tau^f[i]\in[d_1,d_2]\big)}
\Big(\mathbf{m}(S,\vec{x},\varphi_2^{c_2},t,\tau[i])\otimes\bigotimes_{j<i}\mathbf{m}(S,\vec{x},\varphi_1^{c_1},t,\tau[j])\ \Big)\ \Big]
\end{align*}
\begin{align*}
\hspace{-3.5cm}\textbf{C} &\textbf{olored Escape:}\qquad
\mathbf{m}(S,\vec{x},\escape_{[d_1,d_2]}^{f}\varphi^c,t,\ell) = \bigoplus_{\tau\in CRoutes(S(t),\ell,c)}\\
&\Big[\bigoplus_{\ell'\in\tau:\left(d_{S(t)}^f[\ell,\ell']\in[d_1,d_2]\right)}\Big( \bigotimes_{i\le\tau(\ell')}\mathbf{m}(S,\vec{x},\varphi^c,t,\tau[i])\ \Big)\ \Big]
\end{align*}
where $CRoutes(S(t),\ell,c_1,c_2)$ is a message passing function that returns all paths such that every visited location has a color in $c_1$ and terminates with a color in $c_2$, whereas  $CRoutes(S(t),\ell,c)$ returns all paths such that every visited location is in $c$. 
The derived spatial operators inherit the coloring scheme from the reach and escape defined above. More precisely:

\noindent\textbf{Colored somewhere}: $\sw^f_{[0,d]}\varphi^c:=true^C\reach^f_{[0,d]}\varphi^c$;

\noindent\textbf{Colored everywhere}: $\ew^f_{[0,d]}\varphi^c:=\neg\sw^f_{[0,d]}\neg\varphi^c$;

\noindent\textbf{Colored surround}:
$\varphi_1^{c_1}\surr^f_{[0,d]}\varphi_2^{c_2}:= \varphi_1^{c_1}\land \neg\big(\varphi_1^{c_1}\reach_{[0,d]}(\neg(\varphi_1^{c_1}\lor\varphi_2^{c_2}))\big)\land\neg(\escape_{[d,\infty]}\varphi_1^{c_1})$

\begin{example}
    The STREL formula presented in Example~\ref{ex:running1}, Eq.~\eqref{eq:runn1}, 
    describing a scene where a motorbike finds a pedestrian in front within the safety distance, can be rewritten as a {\cSTREL} formula:
    \begin{equation}\label{eq:runn2}
    \varphi = \ev_{[0,T]}\big[ \big(\text{Mov}^{[\mathtt{motorbike}]}(\vec{x})\big)\reach_{[0,d_{\it safe}]}^{Front}\big(\text{Mov}^{\mathtt{[pedestrian]}}(\vec{x})\big)\big],
\end{equation}
where $\mathbf{v}$ is the agent velocity and $\text{Mov}^{[\mathtt{types}]}(\vec{x}) = (\mathbf{v}^{[\mathtt{types}]}(\vec{x}) > 0)$ is the atomic predicated determining whether agents of type $[\mathtt{types}]$ are moving. % or not. 
The robustness of this {\cSTREL} formula quantifies the level of satisfaction\camera{---or criticality---based on the velocities of the motorbike and pedestrian and the proximity of their trajectories.} %, a.k.a. the level of criticality given the velocities of both the motorbike and the pedestrian and the consequent proximity of their trajectories.
\end{example}

\begin{remark}
\camera{We developed a fully differentiable PyTorch library for CSTREL specifications that also provides the first differentiable implementation of STREL, enabling STREL-based gradient searches.}
%We developed a fully differentiable PyTorch library for monitoring CSTREL specifications. As CSTREL generalizes STREL, this library also provides the first differentiable implementation of STREL monitoring, making it applicable to any STREL-based optimization task. %Code will be available upon acceptance.
\end{remark}

\subsection{CSTREL-based Guidance}

Our {\cSTREL} library allows for the flexible, scenario-specific definition and evaluation of formulae describing the targeted behavior. Given a {\cSTREL} formula $\varphi$ and a point $z \in \mathcal{Z}$, the latent space of a pre-trained generative model $G_{\theta}$, we compute a quantitative notion of \textit{robustness} for the corresponding generated trajectory $\tilde{x}^{*} = G_{\theta}(z^*)$ in the data space $\mathcal{X}$. This trajectory $\tilde{x}^{*}$ is a spatio-temporal signal describing the future evolution of the entire scene (all $n=|L|$ agents), conditioned on past observations. 

The {\cSTREL} robustness evaluated at time $0$ is generally a tensor of size $n$. We must therefore find a statistic that effectively summarizes the robustness values in this tensor. In our safety-critical guidance pipeline---where properties describe dangerous behaviors---we aim to find latent inputs that lead to at least one safety violation. The maximum %\footnote{The softmax allows better flow of the gradient, avoiding instabilities.} 
\camera{(the softmax for a better gradient flow)} of the $n$ robustness values is therefore a suitable metric to maximize. Conversely, for realism properties that all agents must satisfy, the minimum value is the appropriate metric to maximize.

Let $\rho^{\varphi}(\tilde{x}^{*})$ denote this real-valued metric, which quantifies the level of satisfaction of $\varphi$ for the scene $\tilde{x}^{}$. Consequently, $\rho^{\varphi}$ serves as a differentiable objective for identifying latent points in $\mathcal{Z}$ that maximize property satisfaction. This guidance effectively forces either at least one agent to violate a safety property or all agents to behave realistically. The two requirements can be combined. %function over all the robustness values to return an aggregated and scalar value while keeping numerical stability.} 

This \cSTREL-based guidance strategy is straightforward to implement but may drive the search toward regions of the latent space that lead to scenes that lie off the data manifold~\cite{wallace2023doodl, karunratanakul2024dno}, i.e. latent inputs with low probability w.r.t. the latent prior $p(z) =\mathcal{N}(0,I)$. To mitigate this issue, we regularize the objective function of our gradient-ascent procedure. More precisely, we jointly optimize over an objective composed of the robustness $\rho^{\varphi}$ and a regularization term proportional to the log-likelihood of the latent input w.r.t the latent prior, which is a standard Gaussian. Mathematically, the regularized objective function is
\begin{equation}\label{eq:ga_loss}
\mathcal{J}(z) = \rho^{\varphi}(G_{\theta}(z)) - \lambda\cdot\left(\frac{1}{2}\|z\|_2^2\right),
\end{equation}
where $\lambda$ is a tunable trade-off parameter. The search can terminate when we find $z^*$ such that $\rho^{\varphi}(G_{\theta}(z^*))>0$. Incorporating realism within the guidance properties can further discourage off-manifold behaviors or hallucinated trajectories. The overall pipeline of our method is summarized in Algorithm~\ref{alg:main_alg}.

%Finally, the expressivity of STREL semantics allows us to incorporate within the logical properties additional \textit{realism checks}, which can further discourage off-manifold behaviors or hallucinated trajectories. 

\begin{algorithm}[!t]
\caption{\textit{STRELGen}}\label{alg:main_alg}
\begin{algorithmic}
    \State \textit{Input:} Diffusion Model $G_{\theta}$, {\cSTREL} formula $\varphi$, starting latent point $z_0 \sim N(\vec{0}, I)$, learning rate $\eta$, regularization parameter $\lambda$.

    \State \textit{Output:} Optimized latent point $z_{\varphi}$.

    \For{$i=1:\text{Max}_{step}$}
        \State $\mathcal{J}(z) = \rho^{\varphi}(G_{\theta}(z)) - \lambda \tfrac{1}{2}\|z\|_2^2,$ \Comment{\texttt{optimization objective}}
        \State $z \gets$ \textbf{GA}$(z,\mathcal{J}(z), \eta)$ \Comment{\texttt{one-step of gradient-ascent}}
    \EndFor

    \If{$\rho^{\varphi}(G_{\theta}(z_{\text{Max}_{step}})) >0$} \Comment{\texttt{formula is satisfied}}
        \State $z_{\varphi} = z_{\text{Max}_{step}}$
    \Else
        \State draw new sample $z^*_0 \sim N(\vec{0}, I)$ and \textbf{go to} step 1 
    \EndIf

\end{algorithmic}
\end{algorithm}

%%%%%%%%%%%%%%%%%%%%%%%%%%%%%%%%%%%%%%%%%%%%%%%%%%%%%%%%%%%%%%%%%%%%%%%%

\section{Experiments}

This section evaluates the proposed approach and addresses the following research questions:
\begin{description}
    \item[\textbf{\textit{RQ1}}] \textit{Can {\cSTREL} guidance reliably steer the diffusion model to produce trajectories that maximize the satisfaction of target specifications?}
    \item[\textbf{\textit{RQ2}}] \textit{Can we guide the model toward safety-critical scenarios while remaining on the data manifold, avoiding low-probability regions of the latent space?}
    \item[\textbf{\textit{RQ3}}] \textit{Can \cSTREL-based guidance optimize latent variables while preserving the physical plausibility of generated trajectories?}
\end{description}
We first describe the experimental setup, followed by qualitative and quantitative analyses of the results. 
\camera{Our implementation is available at \url{https://github.com/lorenzobonin/strelgen}.}

\subsection{Experimental Setup}

\paragraph{Dataset.}
We evaluate our approach on the Argoverse~2 Motion Forecasting Dataset~\cite{wilson2023_argoverse2}, a large-scale benchmark for controllable trajectory generation. 
The dataset comprises over $250{,}000$ driving scenarios collected across six diverse U.S.\ regions, totaling approximately $763$ hours of real-world driving data. 
Each scenario includes a $5$-second observation window followed by a $6$-second prediction horizon. 
We use the official training and validation splits.

\paragraph{Training Details.}
The latent diffusion model is trained for $64$ epochs using the same hyperparameter configuration as in~\cite{wang2024_optimizingdiffusion}, to which we refer for further details. 
Context embeddings are extracted using the pre-trained QCNet model~\cite{zhou2023_qcnet}. 
We employ the implicit formulation of the diffusion process with $100$ denoising steps.

\paragraph{Computational Costs.}
\camera{CSTREL preserves the theoretical complexity of~\cite{nenzi2022logic} over color-based subgraphs. For simple scenarios, our tensorized GPU implementation reduces the time per gradient ascent step of the iterative algorithm in~\cite{nenzi2022logic} from approximately $8\,\mathrm{s}$ to $0.15\,\mathrm{s}$. In more complex settings, this implementation is crucial to maintaining feasibility and computational efficiency.}

%\paragraph{Reproducibility.} %Code will be made available upon acceptance.

\begin{figure*}[!t]
    \centering
    \includegraphics[width=0.75\linewidth]{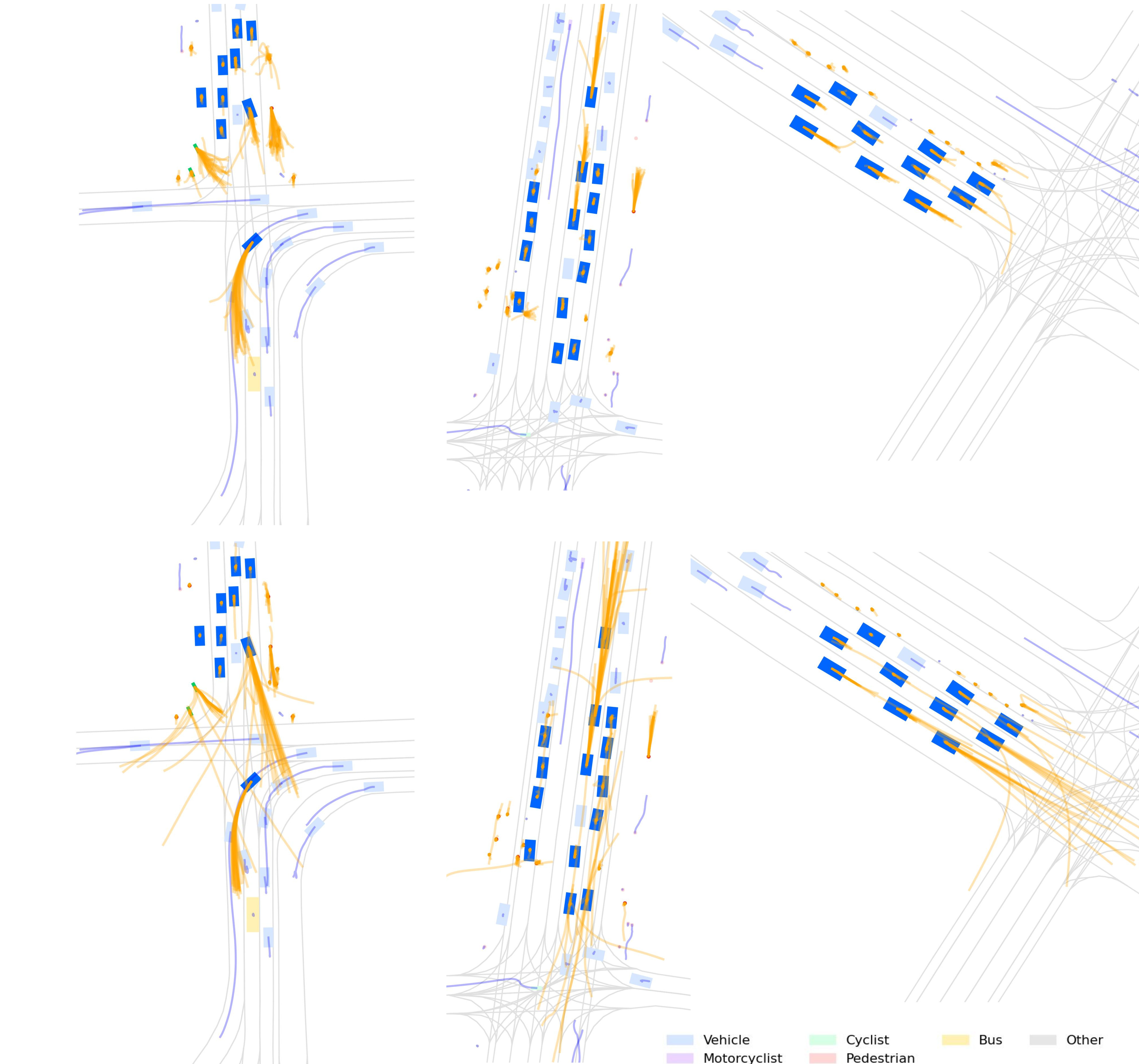}
    \caption{
    Representative safety-critical scenarios generated with \ourmethod. 
    Each column corresponds to a target STREL property: 
    (\textbf{Left}) $\varphi_{pb\_uns}$---\emph{Fast Vehicle Reaches a Bike or Pedestrian};
    (\textbf{Center}) $\varphi_{front}$---\emph{Fast Vehicle Finds a Slow Vehicle Ahead};
    (\textbf{Right}) $\varphi_{surr}$---\emph{Fast Car Surrounded by Slow Cars}.
    Blue trajectories correspond to fixed context agents, and yellow trajectories to optimized predictions.
    The top row shows unguided (vanilla) samples; the bottom row shows samples optimized via \ourmethod.
    }
    
    \label{fig:strelgen_ex}
\end{figure*}

\subsection{{\cSTREL} Specifications}
For evaluation, we selected $n_{\text{scen}} = 3$ representative scenarios from the validation set. 
Specifically, we first identified the ten scenarios containing the largest number of agents and the ten scenarios exhibiting the greatest diversity of agent types. 
From their union, we sampled three scenarios for detailed analysis. \camera{The generative model does not assume a designated ego vehicle, as all agents’ trajectories are generated jointly.}

To evaluate the effect of \cSTREL-based guidance, we designed logical formulae that intentionally promote safety-critical or adversarial behaviors, enabling the model to explore challenging conditions. 
The set of colors $C$ we considered corresponds to the type of agents that can populate the scene. In our experiments $\mathbf{\mathcal{C}} = \{\mathtt{car, pedestrian,bike,motorcycle,bus,static}\}$. To improve readability, we group a subset of relevant labels under the set $\mathtt{vehicles} = \{\mathtt{car, motorcycle, bus}\}$. The considered {\cSTREL} specifications are the following.

%\textbf{Bike or Pedestrian Reaches a Moving Vehicle} 
 \textbf{Fast Vehicle Reaches a Bike or Pedestrian}
%$\varphi_{pb\_uns}$ captures 
capturing potential near-collision events involving vulnerable road users:

    \begin{equation*}\label{eq:ped_unsafe}
        \varphi_{pb\_uns} = \ev_{[0,T]}\Big[
            \big(\text{Fast}^{[\mathtt{vehicles}]}(\vec{x})\big)
            \reach_{[0,d_{\text{safe}}]}^{\text{Euclid}}
          \big(\text{Mov}^{[\mathtt{ped,bike}]}(\vec{x})\big)
        \Big].
    \end{equation*}

 \textbf{Fast Vehicle Finds a Slow Vehicle Ahead} %$\varphi_{front}$ targets 
targeting rear-end collision risks:
    \begin{equation*}\label{eq:reach_fast}
        \varphi_{front} = \ev_{[0,T]}\Big[
            \big(\text{Fast}^{[\mathtt{car,bus}]}(\vec{x})\big)
            \reach_{[0,d_{\text{safe}}]}^{\text{Front}}
            \big(\text{Slow}^{[\mathtt{car,bus}]}(\vec{x})\big)
        \Big].
    \end{equation*}

\textbf{Fast Car Surrounded by Slow Cars} %$\varphi_{surr}$ describes 
describing aggressive driving in dense traffic:
    \begin{equation*}\label{eq:surround}
        \varphi_{surr} = \ev_{[0,T]}\Big[
            \big(\text{Fast}^{[\mathtt{car}]}(\vec{x})\big)
            \surr_{[0,d_{\text{safe}}]}^{\text{Euclid}}
            \big(\text{Slow}^{[\mathtt{car}]}(\vec{x})\big)
        \Big].
    \end{equation*}

%Let $\mathbf{v}(\vec{x})$ denote the velocity of each agent. 
The atomic predicates are defined as:
\iffalse
\[
\text{Mov}^{[\mathtt{types}]}(\vec{x}) = (\mathbf{v}^{[\mathtt{types}]}(\vec{x}) > 0), \]
\[
\text{Fast}^{[\mathtt{types}]}(\vec{x}) = (\mathbf{v}^{[\mathtt{types}]}(\vec{x}) > v_{\text{safe}}), \]
\[
\text{Slow}^{[\mathtt{types}]}(\vec{x}) = (\mathbf{v}^{[\mathtt{types}]}(\vec{x}) < v_{\text{slow}}),
\]
\fi
\camera{$\text{Mov}^{[\mathtt{types}]}(\vec{x}) = (\mathbf{v}^{[\mathtt{types}]}(\vec{x}) > 0) \land (v_{\text{real}} >\mathbf{v}^{[\mathtt{types}]}(\vec{x}))$, $\text{Fast}^{[\mathtt{types}]}(\vec{x}) = (\mathbf{v}^{[\mathtt{types}]}(\vec{x}) > v_{\text{safe}}) \land (v_{\text{real}} >\mathbf{v}^{[\mathtt{types}]}(\vec{x}))$ and $\text{Slow}^{[\mathtt{types}]}(\vec{x}) = (\mathbf{v}^{[\mathtt{types}]}(\vec{x}) < v_{\text{slow}})$, }
where \camera{$\mathbf{v}(\vec{x})$ is the velocity of each agent}, $v_{\text{safe}}$, $v_{\text{real}}$, and $v_{\text{slow}}$ are velocity thresholds and $d_{\text{safe}}$ is the safety distance. 
These parameters are scenario-dependent (e.g., thresholds differ between urban and highway scenes). 
Each specification is applied to one of the selected scenarios.
\camera{To further demonstrate the versatility of CSTREL, we introduce two additional formulas that, although not safety-critical, promote realistic motion and smooth behavior.}
%To further highlight the versatility of the {\cSTREL} framework, we additionally define two formulae that, although outside the safety-critical generation scope, enforce realistic motion patterns and smooth behavior.

\textbf{No Sudden Changes in Heading Direction:}
    \begin{equation*}\label{eq:globhead}
        \varphi_{\text{head}}
        =
        \glob_{[0,T]}
        \big( \mathbf{h}^{[\mathtt{car}]}(\vec{x}) < h_{\text{smooth}} \big),
    \end{equation*}
 \begin{align*}\label{eq:everywhere_safe}
     \textbf{No }&\textbf{Overlap Among Moving Vehicles:}\ \varphi_{\text{ov}}=\glob_{[0,T]}
        \nonumber \\
        &\ew^{\text{Euclid}}_{[0,D]}\Big(\neg\big(
                \text{Mov}^{[\mathtt{vehicle}]}(\vec{x})
                \land
                \sw^{\text{Euclid}}_{(0,\,d_{\text{safe}}]}
                \text{Mov}^{[\mathtt{vehicle}]}(\vec{x})
            \big)
        \Big),
    \end{align*}
where $\mathbf{h}(\vec{x})$ denotes the instantaneous heading change, and $D$ and $d_{\text{safe}}$ are distance hyperparameters.

\paragraph{Baseline.}
As a baseline, we use the vanilla diffusion model without any \cSTREL-based guidance. This comparison isolates the contribution of the proposed method by contrasting guided and unguided generations under identical conditions.

\paragraph{Metrics.}
Performance is assessed through both qualitative and quantitative analyses. 
Qualitatively, we inspect generated trajectories to evaluate plausibility, diversity, and interpretability (%shown in 
Fig.~\ref{fig:strelgen_ex}). 
Quantitatively, we compute the distribution of minimum pairwise distances $D$ between agents (%shown in 
Fig.~\ref{fig:boxplot}) and the number of collisions observed in each scenario (%shown in 
Table~\ref{tab:collisions}). Together, these metrics provide complementary insights into how {\ourmethod} 
steers the diffusion model. %while maintaining realism.

\subsection{Results}
Fig.~\ref{fig:strelgen_ex} shows representative examples of safety-critical scenes generated via {\ourmethod}. 
%The unguided (vanilla) samples are shown in the top row, while optimized trajectories appear in the bottom row. 
\camera{The top row shows unguided (vanilla) samples, while the bottom row presents optimized trajectories.}
Across all three properties, the optimized trajectories exhibit behaviors consistent %aligned
with the target logical formulae: pedestrians and cyclists approach vehicles (\textbf{left}); fast vehicles approach slower ones ahead (\textbf{center}); and aggressive drivers emerge within congested traffic (\textbf{right}). 
In all cases, trajectories remain spatially coherent and aligned with the road geometry.
Importantly, {\cSTREL} ensures that only the agent types %referenced in each specification 
\camera{specified in each formula} exhibit significant behavioral changes relative to the baseline. 
The likelihood regularization term %$\lambda$
\camera{in Eq.~\eqref{eq:ga_loss}} prevents the optimization from deviating toward low-probability regions of the latent space. 
We also considered composing safety-critical and other realism-oriented formulae (e.g., Eq.~\eqref{eq:globhead}), but empirically found that using likelihood penalty was sufficient to obtain plausible trajectories.

\begin{figure}[!t]
    \centering
    \includegraphics[width=.9\columnwidth]{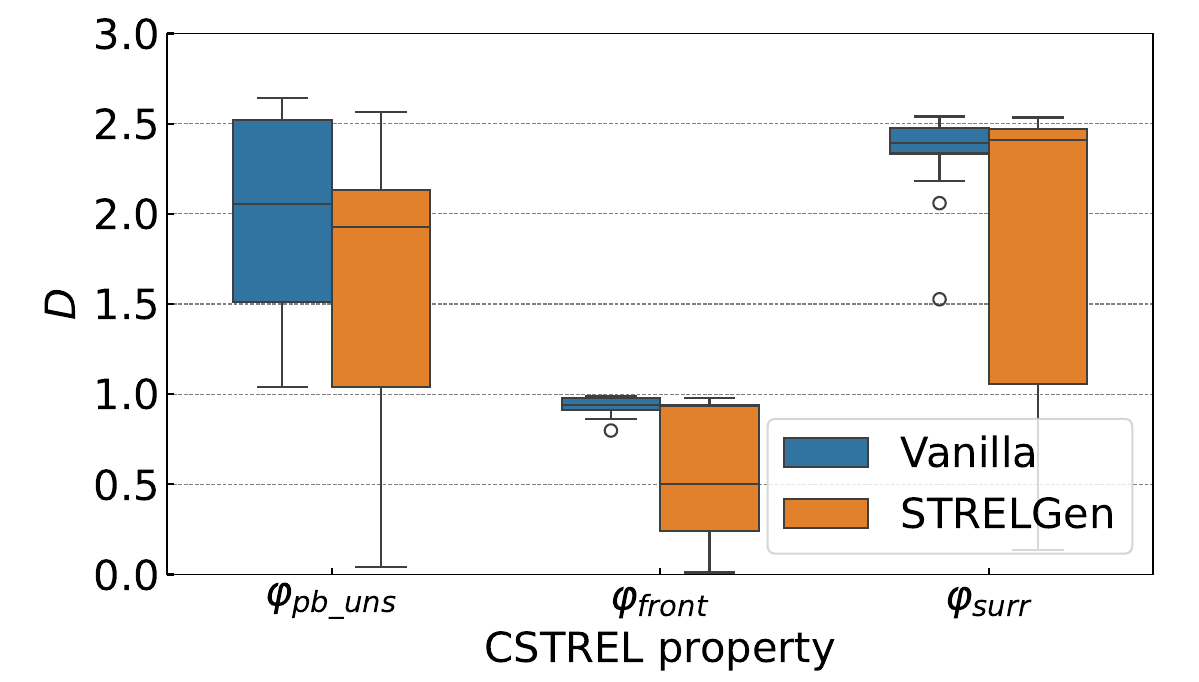}
    \caption{
    Distribution of minimum inter-agent distance $D$ (in meters) across {\cSTREL} properties\camera{; lower values indicate closer agents and thus higher safety risk.}%.     Lower values correspond to smaller separations between agents and thus higher safety risk.
    } \label{fig:boxplot}
   
%\end{figure}

%\begin{table}[!b]
%\centering
\vspace{3mm}
\begin{tabular}{lccc}
\hline
\textbf{Method / Property} & $\varphi_{pb\_uns}$ & $\varphi_{front}$ & $\varphi_{surr}$ \\
\hline
\emph{Vanilla} & 0 (0\%) & 5 (17\%) & 0 (0\%) \\
\ourmethod & 5(17\%)& 19 (63\%)& 7 (23\%)\\
\hline
\end{tabular}
\caption{Potential collision: count and percentage %by configuration 
(30 generated scenarios for each method-formula configuration). %Based on 20 generated scenarios per method and cSTREL property.
%%%Counts of scenarios with at least one collision across methods and {\cSTREL} properties over 20 scenarios generated for each configuration and relative percentage.
}
\label{tab:collisions}
\end{figure}

To quantify the steering effect of {\ourmethod}, Fig.~\ref{fig:boxplot} reports the distribution of the minimum inter-agent distance $D$---measured between the center points of the agents---across {\cSTREL} properties, comparing the baseline (blue) and guided (orange) generations. 
{\ourmethod} consistently produces lower median values of $D$, indicating reduced safety margins and a higher likelihood of unsafe interactions. 
The method also exhibits larger interquartile ranges, reflecting increased diversity in the generated scenarios.
\camera{With respect to the considered experiments, the proportion of generated scenarios that are challenging, meaning positive robustness w.r.t. the CSTREL requirements, is around 13,6\% under unguided sampling, increasing to 100\% with the proposed CSTREL-based guidance strategy.}
Table~\ref{tab:collisions} reports the number of collisions observed per {\cSTREL} property. \camera{We define a potential collision as a vehicle being within 0.9 m of another agent.} %We define a potential collision between \camera{a vehicle and any other type of agent if the distance between their center points is less than $0.9$ meters.} 
%two vehicles as occurring when the distance between their center points is less than $1$ meter, and between a vehicle and any other type of agent when the distance is less than $0.5$ meters.
Across all cases, {\ourmethod}  consistently increases the number of potential collisions, confirming its effectiveness in generating safety-critical behaviors while remaining within the data manifold.

%\end{table}

\section{Discussion}
In this section, we summarize the key findings that address the research questions posed in this work.

\begin{description}
    \item[\textbf{\textit{RQ1}}] \textit{Can {\cSTREL}  guidance reliably steer the diffusion model to produce trajectories that maximize the satisfaction of target specifications?}

     \camera{The results section shows that} %As demonstrated in the results section, 
     the proposed approach effectively generates safety-critical scenarios that satisfy the properties specified through {\cSTREL}, offering explicit control over both global scene behavior and agent-specific dynamics. The integration of {\cSTREL} guidance steers the diffusion process toward configurations that maximize the desired specifications, enabling targeted manipulation of safety margins and interaction patterns among agents.

     \item[\textbf{\textit{RQ2}}] \textit{Can we guide the model toward safety-critical scenarios while remaining on the data manifold, avoiding low-probability regions of the latent space? }

     % The guidance with robustness alone can lead to latent points that lie in the tails of the latent distribution, to which corresponds generated samples that are not in the data manifold. To avoid going off-manifold, we integrated in our robustness function a likelihood penalty. 
     % While rather simple, with this inclusion we found that the optimized latent points that {\ourmethod} produces are realistic with respect to the latent distribution, while still leading to output that maximize the target property.

     Guidance based solely on robustness can push latent points toward the tails of the distribution, producing off-manifold samples. To mitigate this, we incorporated a likelihood penalty into the robustness function. This simple yet effective addition ensures that the optimized latent points generated by {\ourmethod} remain consistent with the latent distribution, while achieving high satisfaction values for the target property.

    \item[\textbf{\textit{RQ3}}] \textit{Can \cSTREL-based guidance optimize latent variables while preserving the physical plausibility of generated trajectories?}

    % We considered the possibility of including realism-preserving formulas in our robustness to avoid hallucinations, but we found out that the likelihood penalty was enough to keep also the generated trajectory realistic. 
    % This shows that we can use {\ourmethod} not only to perform scenario generation, but also to explore the coverage of the underline generative model with respect to the desired properties by testing its ability to produce compliant samples while staying in distribution.

    We explored the inclusion of realism-preserving formulae in the robustness function but found that the likelihood penalty alone was sufficient to obtain plausible trajectories. In fact, our qualitative analysis showed that the generated samples tend to adhere to the underlying physical and environmental constraints.
    This result indicates that {\ourmethod} not only supports targeted scenario generation but also enables analysis of the generative model’s coverage with respect to desired properties—revealing its capacity to produce compliant samples while remaining within the data manifold.
    
\end{description}
Overall, our findings suggest that {\ourmethod} can successfully balance specification-driven optimization with the plausibility of generated samples, indicating its potential for controllable and interpretable scenario generation in safety-critical domains.

%%%%%%%%%%%%%%%%%%%%%%%%%%%%%%%%%%%%%%%%%%%%%%%%%%%%%%%%%%%%%%%%%%%%%%%%

\section{Conclusions}

In this work, we introduced {\ourmethod} a controllable trajectory generation framework based on Colored Signal Spatio-Temporal Reach and Escape Logic (\ColSTREL), an extension of STREL that enables the specification and quantitative monitoring of spatio-temporal properties across different classes of agents. By leveraging {\ColSTREL} robustness as a differentiable objective, we guide a pre-trained diffusion model in its latent space to generate trajectories that satisfy complex, formally defined safety-critical behaviors.

The proposed guidance strategy, which combines the {\ColSTREL} objective with a likelihood-based regularization term, proved effective in producing safety-critical scenarios. The regularization term ensures that the optimization remains close to the data manifold, while the logical objective directs the generation toward behaviors exhibiting reduced safety margins or other targeted interactions. Overall, these results demonstrate that formal, logic-based guidance can complement data-driven generative models, bridging the gap between realism and controllable scenario synthesis for simulation-based safety testing.

Future work will aim at developing automated pipelines that infer safety-critical {\cSTREL} specifications directly from the underlying road geometry and agent configurations, thereby removing the need for manual formula design and enabling targeted optimization toward context-dependent unsafe behaviors.
In addition, the proposed framework could be extended to evaluate autonomous driving policies under adversarial or rare-event conditions by optimizing latent variables adversarially with respect to the control policy of interest.

%%%%%%%%%%%%%%%%%%%%%%%%%%%%%%%%%%%%%%%%%%%%%%%%%%%%%%%%%%%%%%%%%%%%%%%%

%%%%%%%%%%%%%%%%%%%%%%%%%%%%%%%%%%%%%%%%%%%%%%%%%%%%%%%%%%%%%%%%%%%%%%%%
% Acknowledgments
%%%%%%%%%%%%%%%%%%%%%%%%%%%%%%%%%%%%%%%%%%%%%%%%%%%%%%%%%%%%%%%%%%%%%%%%

\section*{Acknowledgments}

This work has been partially supported by the iNEST project funded by the European Union Next-GenerationEU (PNRR -- Missione 4 Componente 2, Investimento 1.5 -- D.D. 1058 23/06/2022, ECS\_00000043) and by the National Science Foundation through the following grants: CAREER award (SHF-2048094), CNS-2039087, IIS-SLES-2417075, and funding by Toyota R\&D through the USC Center for Autonomy and AI. This work does not reflect the views or positions of any organization listed.

%%%%%%%%%%%%%%%%%%%%%%%%%%%%%%%%%%%%%%%%%%%%%%%%%%%%%%%%%%%%%%%%%%%%%%%%
% Bibliography
%%%%%%%%%%%%%%%%%%%%%%%%%%%%%%%%%%%%%%%%%%%%%%%%%%%%%%%%%%%%%%%%%%%%%%%%

\bibliographystyle{plainnat}
\bibliography{references}

%%%%%%%%%%%%%%%%%%%%%%%%%%%%%%%%%%%%%%%%%%%%%%%%%%%%%%%%%%%%%%%%%%%%%%%%

\end{document}